\documentclass[11pt]{article}

\usepackage[T1]{fontenc}
\usepackage{lmodern}
\usepackage[margin=1in]{geometry}
\usepackage{graphicx}
\usepackage{amsmath}
\usepackage{amssymb}
\usepackage{booktabs}
\usepackage{multirow}
\usepackage{caption}
\usepackage{subcaption}
\usepackage[hidelinks]{hyperref}
\usepackage{graphicx}
\graphicspath{{pictures/}}

\usepackage{fancyhdr}
\begin{document} 

\title{Topological Fraud Detection in Latent Transaction Spaces}
 \author{Avraham Bourla}
 \date{\today}
 \maketitle

\begin{abstract}
\noindent Working entirely on topologically anonymized embeddings, we perform fraud detection using iterative rounds of unsupervised filtering followed by supervised sniping. The result is an ultra-low latency privacy--preserving triage that allows institutions to flag suspicious activity without compromising Personally Identifiable Information.
\end{abstract}

\section{Introduction}

\subsection{The Privacy Preserving Real Time Fraud Detection Challenge}

\noindent Although state-of-the-art (SOTA) fraud detection continues to reach new performance benchmarks, the global arms race is intensifying as adversarial tactics evolve to exploit the remaining blind spots in institutional defenses. Global fraud losses are on a steep upward trajectory, accelerating toward a projected \$362 billion—a staggering increase from the estimates of just a few years ago. This surge is reflected in the mounting economic drain per incident; while a single fraudulent dollar cost merchants \$3.36 in 2020, that figure has now climbed to \$4.23 \cite{LexisNexis}. Because privacy regulations and competitive barriers prevent the sharing of raw transactional data, criminals can exploit the widening information gaps within "black box" institutional security to distribute adversarial patterns across multiple entities without triggering localized detection thresholds. This disconnected defensive landscape ensures that fraudsters scale their operations significantly faster than formal data-sharing agreements can evolve \cite{FATF}. \\

\noindent Despite the widespread recognition that sharing transactional signatures would drastically enhance collective detection rates, titans like Visa and Mastercard continue to hoard data in order to protect their proprietary 'moats' and maintain a strategic edge over competitors \cite{Mastercard, Visa}. This defensive isolation leaves the rest of the ecosystem—including banks, merchants, and emerging fintech players—at a severe loss, trapped within information silos with only binary labels and fragmented datasets to inform their risk assessments. This data asymmetry creates a critical mandate for privacy-preserving fraud detection models that can facilitate cross-institutional intelligence sharing without compromising competitive advantages or consumer confidentiality.\\

\noindent While Fully Homomorphic Encryption (FHE), which enables direct computation on ciphertexts, represents an ideal theoretical framework for privacy-preserving networks, it suffers from notoriously high latency, often measured in seconds or even minutes. Such a delay is fundamentally incompatible with the demands of real-time commerce, where sub-millisecond inference has emerged as the gold-standard benchmark for the next generation of transactional AI. Beyond enabling the rapid interception of overt threats and the line-rate filtering of legitimate traffic, such ultra-low latency preserves a frictionless user experience while fortifying the system against high-volume automated attacks \cite{Zhu}. The private sector offers more viable solutions such as Privacy-Enhancing Technologies (PETs) and multi-party secret-sharing protocols \cite{Gupta2025}. Nevertheless, these cryptographic primitives face significant latency bottlenecks and communication overheads, with the bottom line being a far cry from the desired 1ms benchmark.\\

\noindent \noindent Achieving ultra-low-latency, privacy-preserving inference will not only level the playing field for smaller fintech entities but also propel the financial sector toward the collaborative maturity seen in medicine and genomics, where pooling privacy-protected insights enables the identification of systemic anomalies \cite{Torkamani}. Under this paradigm, a single institution's detection of a fraudulent pattern can trigger a real-time, network-wide warning, shielding participating institutions before the threat spreads. This decentralized financial "immune system" embodies Bruce Schneier's observation that "security is a process, not a product" \cite{Schneier}. Specifically, an institution can broadcast an anonymized fraud signature to participating institutions, where the model identifies similar transactions and flags them as hot for immediate scrutiny. By propagating these warnings with ultra-low latency, the network could enable earlier detection of coordinated fraud rings across institutional boundaries, potentially limiting their reach and reducing resulting losses.
 
\subsection{Topological Methods in Anomaly Detection}

\noindent Current regulatory frameworks permit financial institutions to share anonymized datasets by decoupling sensitive identifiers from behavioral signatures \cite{EU}. Under these frameworks, participants can exchange mathematical fingerprints that capture structural correlations that are virtually impossible to spoof with simple aliases. The semantic labels of specific features—whether a field represents a `home address' or `V1'—are irrelevant in the latent space; only the mathematical dissonance and geometric isolation of fraudulent clusters matter. Consequently, a topological approach allows for the identification of fraud through behavioral geometry alone, providing a robust detection layer that is indifferent to the specific methods of data encryption or anonymization \cite{Chandola, Liu}.\\

\noindent Our central hypothesis is that fraudulent activity is localized within distinct, well-formed clusters within latent topological embedding spaces \cite{Zhu}. Empirical observations suggest that Uniform Manifold Approximation and Projection (UMAP) is superior to alternative manifold learning techniques for preserving the structural nuances required to visualize fraudulent transactions \cite{McInnes, umap}. For classification, tree-based models consistently deliver superior performance on the tabular datasets that constitute the vast majority of mission-critical transactional data \cite{Grin}. According to this seminal work, such models provide high-accuracy results in a fraction of the time required by computationally intensive deep learning alternatives, making them the ideal choice for sub-millisecond synchronous gatekeeping.\\

\noindent Two further ideas organize the method. The first is \emph{iteration}: rather than asking a single model to separate a rare class from an overwhelming majority, the majority is peeled away in successive rounds, so that each round faces a smaller and more informative residual. This is the logic of the attentional cascade, in which cheap early stages reject easy negatives \cite{ViolaJones}, and of iterative undersampling schemes that repeatedly discard majority examples already handled well \cite{LiuWuZhou}. The second is \emph{persistent homology}, which summarizes the shape of data by tracking how the connected components (and higher-dimensional holes) of a growing family of sublevel sets are born and merge as a threshold rises \cite{Edelsbrunner, ZomorodianCarlsson, Carlsson}. Features that persist over a wide range of thresholds are structure; short-lived ones are noise, and the resulting summaries are provably stable under small perturbations of the input \cite{CohenSteiner}. For functions sampled on a regular grid, the computation reduces to a cubical complex \cite{Wagner}, which makes it a natural companion to grid-based representations such as the Self-Organizing Map, whose U-matrix of neighboring-prototype distances is itself a function on the grid \cite{UltschSiemon}. When the function changes from round to round, the evolution of its persistence diagram can be followed as a vineyard \cite{Vineyards}. Introductions aimed at data scientists are given in \cite{ChazalMichel}.\\

\noindent While classic tree-based models often outperform deep learning on tabular datasets, the expressive power of the latter is indispensable for the nuanced task of characterizing 'normal' activity \cite{Goel}. We train an autoencoder (AE) exclusively on legitimate behavior, so that it captures the manifold of typical transactions \cite{ZhouPaffenroth, Chalapathy}. Its bottleneck is used in two ways: the distance of a transaction to an unsupervised `hot' cluster in the bottleneck becomes a scalar anomaly signal, and the bottleneck coordinates themselves feed a second map of the data. Both the UMAP latent and the AE latent are organized by Self-Organizing Maps (SOMs) \cite{Kohonen}: toroidal grids of prototype vectors, retrained from scratch at every iteration on the legitimate rows that remain, onto which every transaction is projected. A transaction's distance to its nearest prototype (the quantization error) is a further anomaly signal, and the AE and UMAP signals are blended into a single per-cell safety score. Because the SOM is a grid, it supports exact topological operations—connected components, complements, persistent homology—that are ill-defined on a raw point cloud, and these operations drive the filtering described below \cite{Adejoh, Carlsson}.

\section{Methodological Prerequisites: EDA and Feature Engineering}

\noindent We begin by deploying the Kaggle credit card anonymous fraud dataset \cite{Kaggle}, which is a publicly available, highly imbalanced dataset of actual European card transactions. The features V1 through V28 are anonymized Principal Component Analysis (PCA) transformations of the original confidential variables, preserving privacy while retaining predictive structure. The fraudulent transactions form relatively tight clusters in feature space; Figure 1(a) is a simple scatterplot of the two most important PCA features. One can clearly observe the vast majority of fraudulent transactions lying outside a single dense cluster. While the dataset may appear insufficiently complex for the deployment of a production-grade fraud detection model, it yields the critical exploratory observation: the log base--10 term of the amount has a rough normal distribution.\\

\begin{figure}[h]
    \centering
	\begin{subfigure}{0.48\textwidth}
        \raisebox{.3cm}{%
            \includegraphics[width=\linewidth,height=5.9cm]{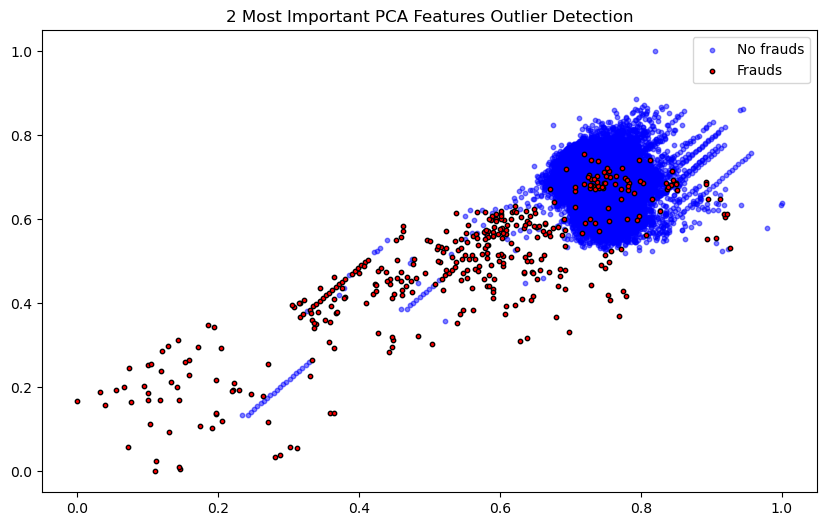}
        }
        \caption{PCA Clusters}
     \end{subfigure}
	\begin{subfigure}{0.48\textwidth}
        \centering
        \includegraphics[width=\linewidth]{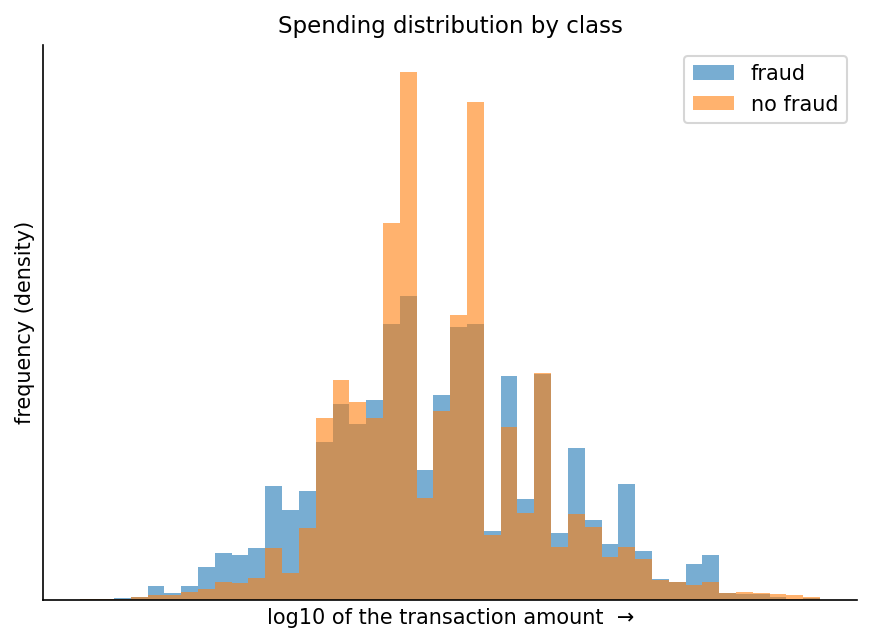}
        \caption{Log-Amount Dist.}
    \end{subfigure}
    \caption{Exploratory Data Analysis of the Kaggle Dataset}
    \label{fig:two_plots}
\end{figure}

\noindent We add the logarithmic z--score feature:
\[ z_{\log} = \frac{\log_{10}(\text{Amount}) - \mu}{\sigma} \]
where $\mu$ and $\sigma$ are the prospective mean and standard deviation. Furthermore, we incorporate the leading significant digit—facilitating an analysis of Benford’s Law—alongside the fractional component of each transaction (00–99 cents). This approach is motivated by the heuristic that the distribution of decimal values in fraudulent transactions often exhibits significant divergence from benign patterns. Consequently, we map the transaction amount $x$ to a synthetic feature space by defining the signature:
\[ f(x) = \operatorname{FirstDigit}(x) + \frac{\operatorname{Cents}(x)}{100}. \]
All splits are time-ordered (a 75/5/20 train/validation/test split), so that every evaluation predicts the future from the past.

\section{An Iterative Topological Sniping/Filtering Scheme}

\noindent We now turn our attention to the IEEE-CIS dataset \cite{IEEE}, a large-scale, real-world e-commerce dataset provided by Vesta Corporation. We perform two independent projections of the imputed, scaled raw data, from which location columns are removed: a UMAP embedding and the AE bottleneck. The UMAP coordinates together with the amount signature form the latent space in which the iterations operate; the AE contributes its own latent coordinates and the distance to its hot cluster. As shown in the depicted plots, these AE-derived signals separate the legitimate and fraudulent rows.\\

\begin{figure}[htbp]
    \centering
    \begin{subfigure}{0.49\textwidth}
        \centering
        \includegraphics[width=\linewidth, height=6cm]{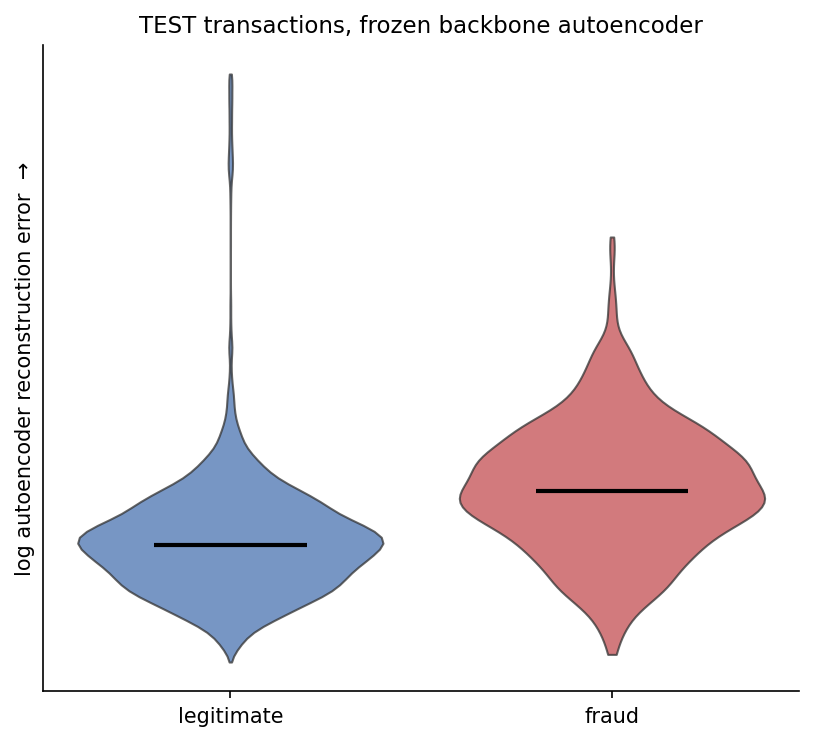}
        \caption{Reconstruction Error}
        \label{fig:reconstruction}
    \end{subfigure}
    \hfill
    \begin{subfigure}{0.49\textwidth}
        \centering
        \includegraphics[width=\linewidth, height=6cm]{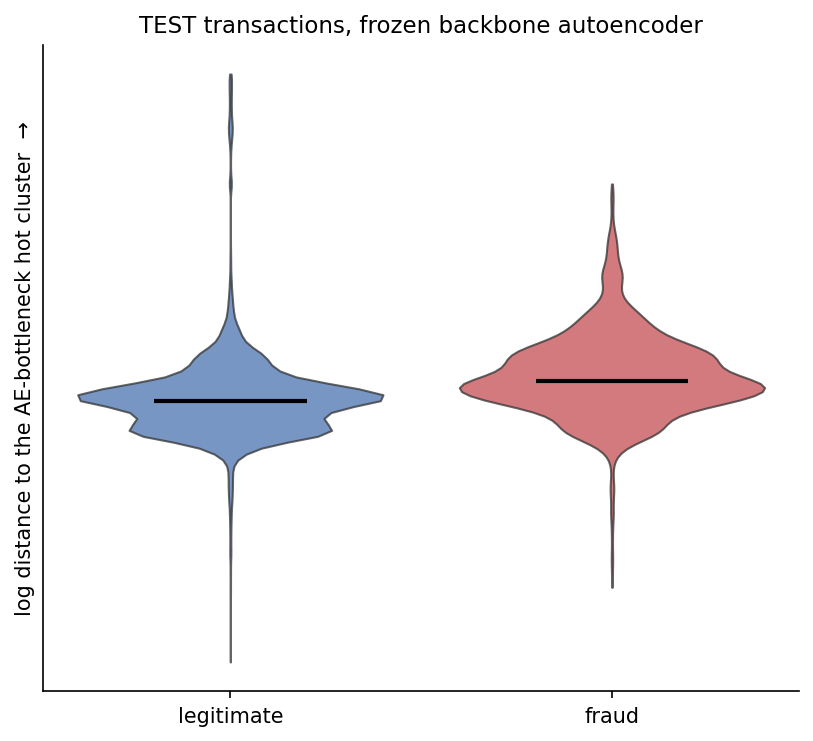}
        \caption{Distance to Centroid}
        \label{fig:umap_clustering}
    \end{subfigure}
    \caption{Comparative analysis of the Autoencoder features: legitimate vs. fraud}
    \label{fig:triage_plots}
\end{figure}

\noindent The backbone and the iterative scheme are illustrated below. Each iteration removes rows it is confident are legitimate (hard removal) and sets aside rows it finds suspicious (soft removal); a single final classifier then scores everything that was not removed.

\noindent\includegraphics[width=\linewidth, height=7cm]{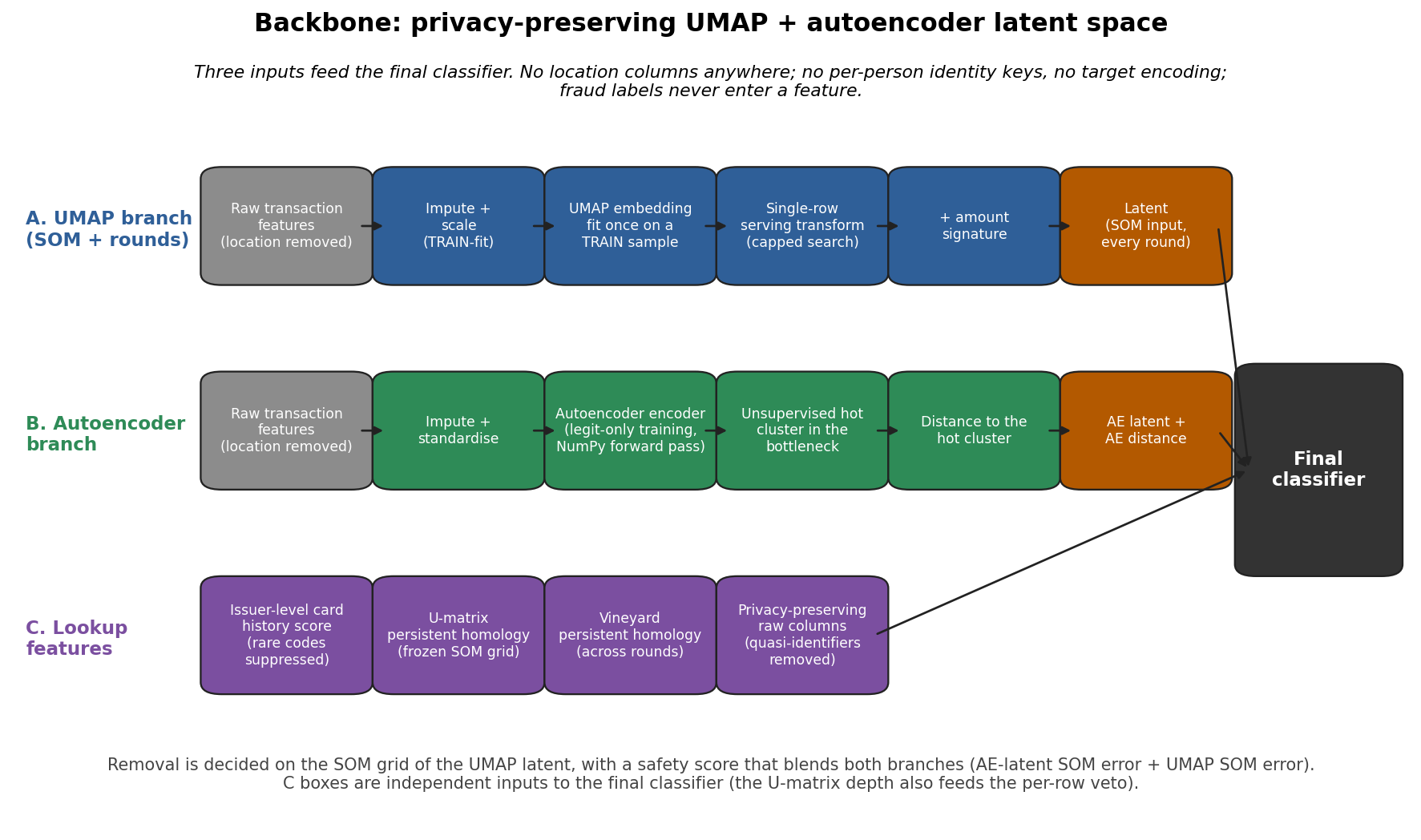}

\noindent\includegraphics[width=\linewidth, height=11cm]{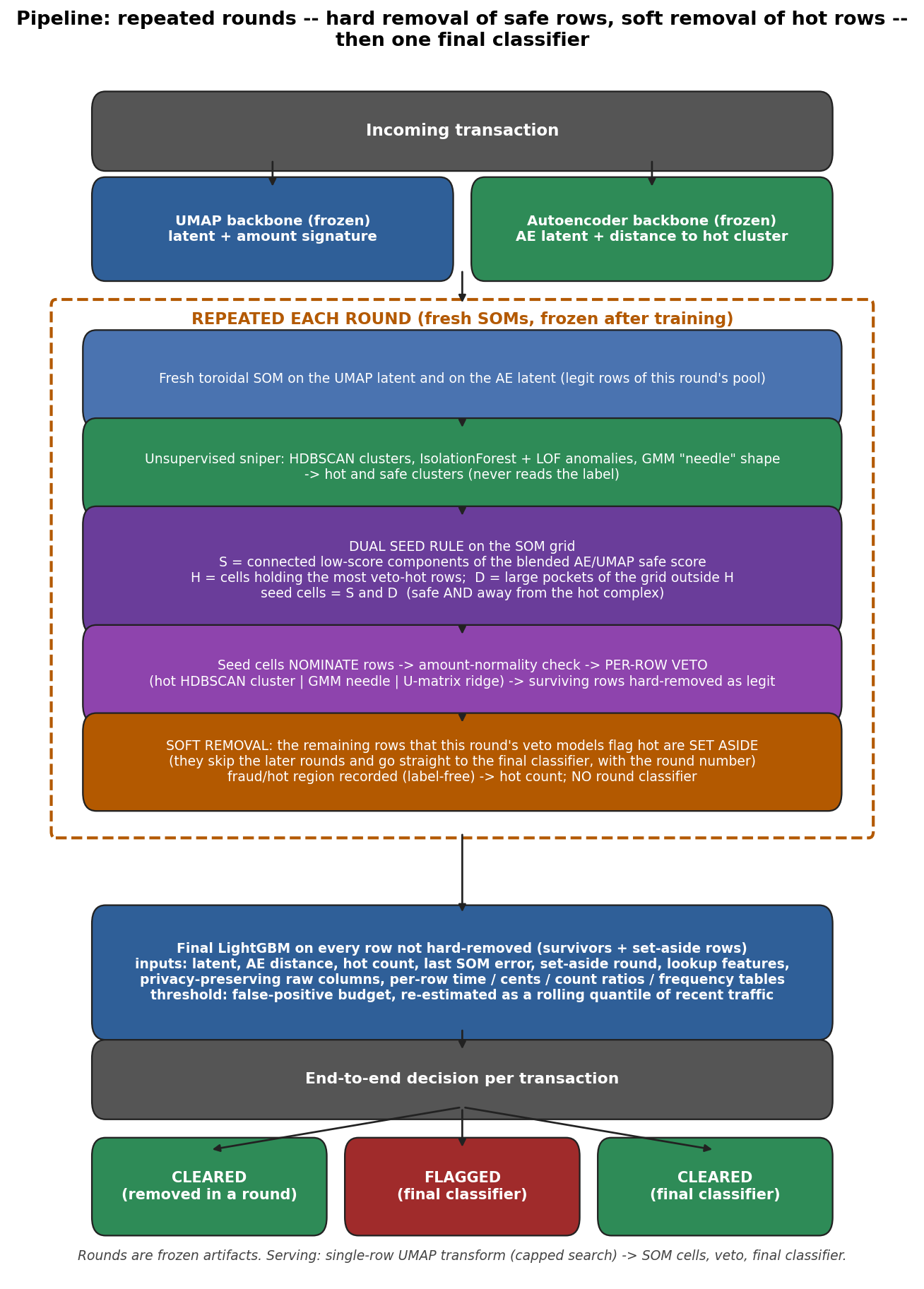}

\subsection{Pre-sniping via Classical Anomaly Detection}

\noindent Effective fraud sniping requires not only working in small batches but also undersampling each batch, retaining only the least dense half of the observations. This enables the emergence of distinct structural patterns that would otherwise be obscured by the preponderance of benign data.\\

\noindent\includegraphics[width=\linewidth, height=9cm]{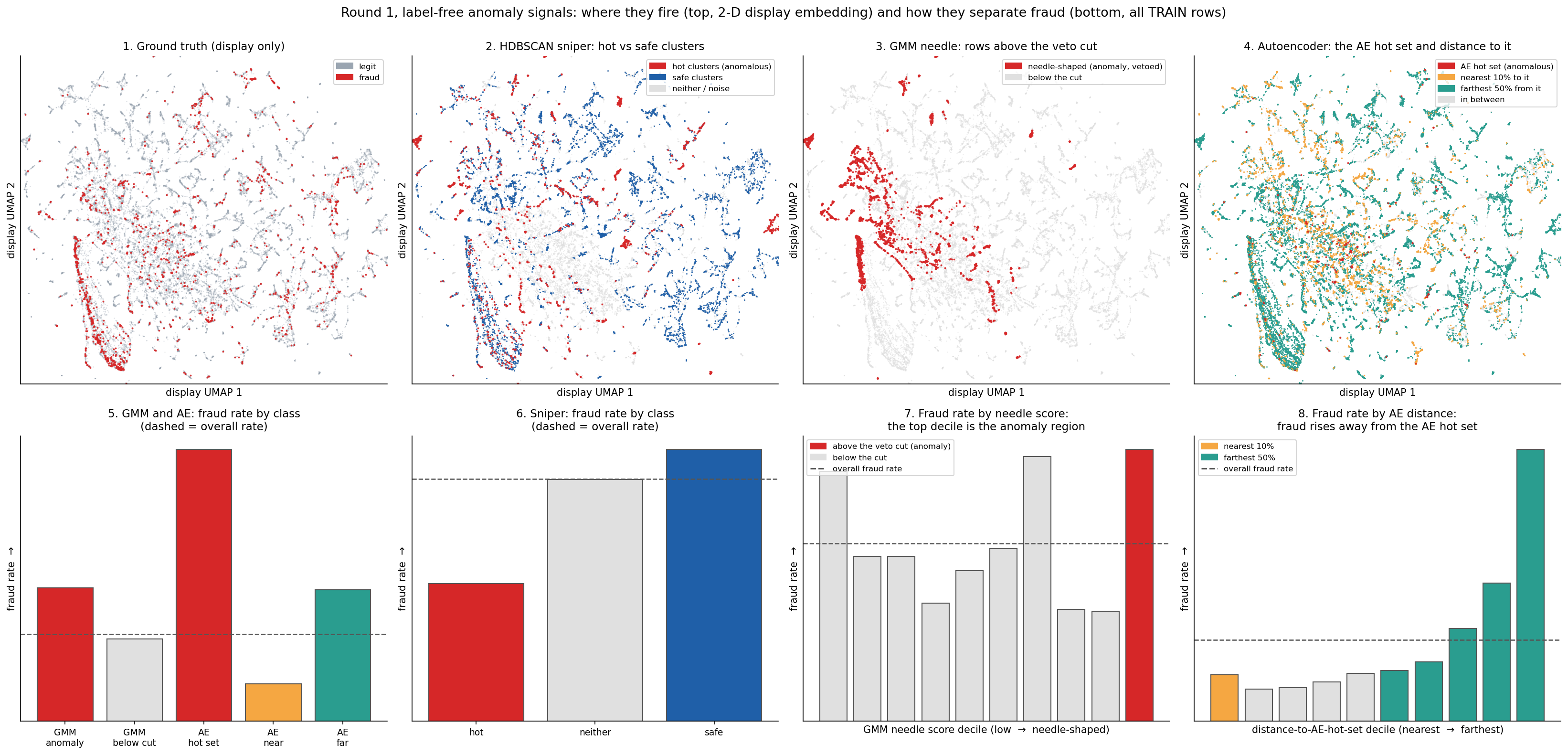}

\noindent The unsupervised UMAP embedding is clustered with HDBSCAN \cite{Campello}, where dense neighborhoods form structural kernels while sparse points are relegated to noise. These clusters are subsequently evaluated for anomalous characteristics using Isolation Forest (IF) \cite{Liu} and the Local Outlier Factor (LOF) \cite{Breunig}. Within the anomaly-flagged subsets, we apply a Gaussian Mixture Model (GMM) to analyze sub-cluster geometry; each component is scored by its covariance eigenvalue ratio—an anisotropic measure of ``eccentricity''—under the heuristic that tight, directional sub-clusters may indicate coordinated fraud \cite{Chandola, gmm}. The labels are never read at this stage. In practice, the GMM eccentricity and the autoencoder's hot set separate fraud far better than the cluster labels themselves, as the figure above shows. Proximity to the AE hot set, moreover, is not a hot score: the hot set itself is strongly enriched in fraud, but its immediate neighborhood is among the least fraudulent parts of the data, and the fraud rate rises with distance from it. The AE distance is therefore passed to the final classifier as an input rather than used as a fixed rule.

\subsection{Filtering on the SOM Grid via the Dual Seed Rule}

\noindent The central innovation of this methodology is the hard removal of legitimate rows, and it is decided entirely on the SOM grid. Two complexes of cells are built from label-free signals. The \emph{safe complex} $S$ consists of the connected components of the low sublevel set of the blended AE/UMAP safety score; isolated cells are discarded. The \emph{hot complex} $H$ collects the cells holding the largest share of rows that the sniper's veto models flag as hot. Safe cells are only trusted where they lie in the \emph{dual} of the hot complex: the complement of $H$ is split into connected regions, and the small pockets that $H$ cuts off are excluded, leaving $D$. Following the standard pairing of digital topology, $S$ is taken with 4-connectivity and the complement of $H$ with 8-connectivity \cite{KongRosenfeld}. The seed cells are $S \cap D$: safe, and not enclosed by the hot complex.\\

\noindent\includegraphics[width=\linewidth, height=10cm]{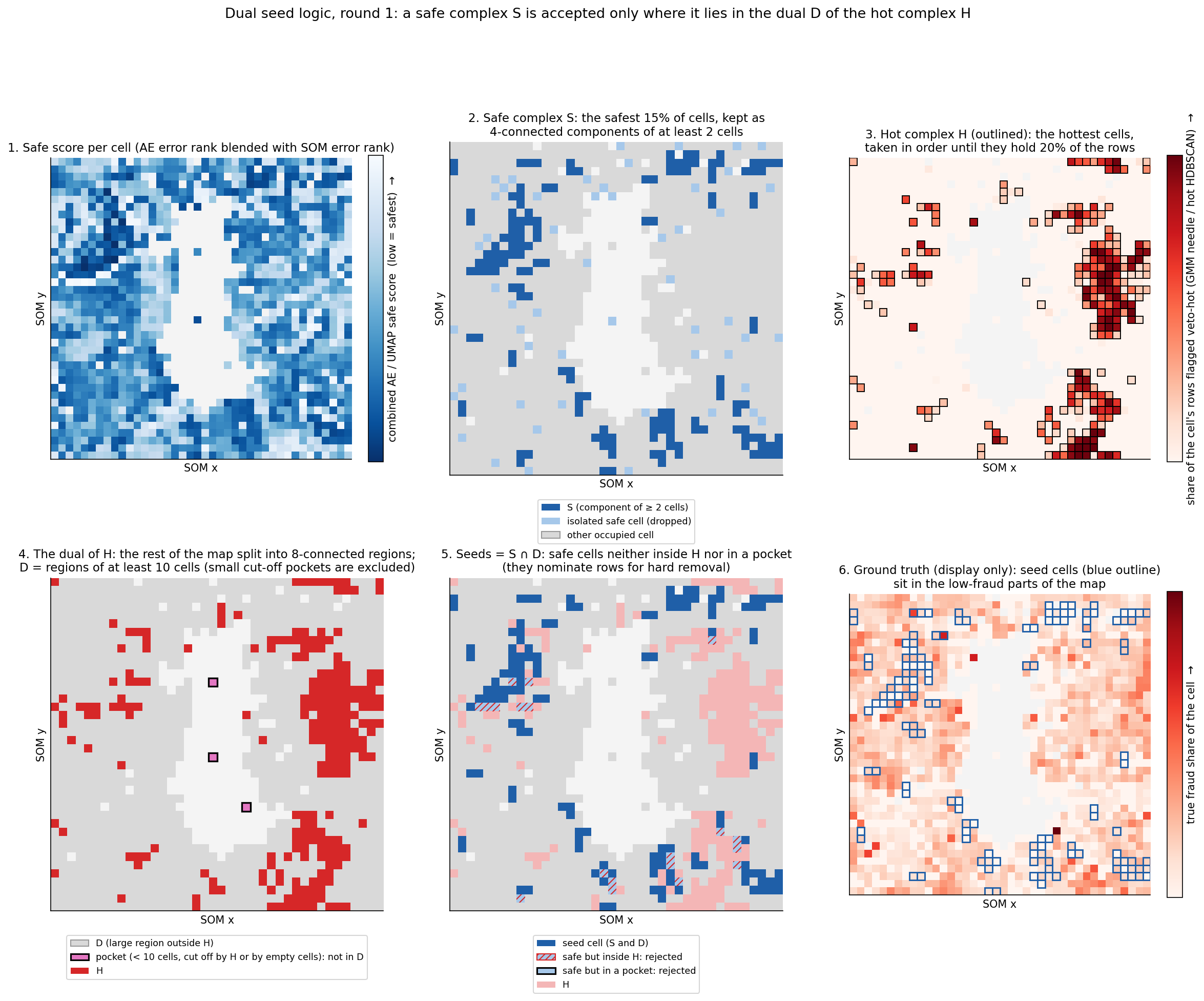}

\noindent Seed cells only \emph{nominate} rows. A nominated row is hard-removed as legitimate only if it also passes an amount-normality check and none of the per-row vetoes fires: membership of a hot HDBSCAN cluster, a high GMM eccentricity, or a location on a ridge of the SOM landscape. The last check is topological. On one frozen SOM we compute the sublevel-set filtration of the U-matrix on the torus and its zero-dimensional persistent homology \cite{Edelsbrunner, Wagner}: basins of the U-matrix are dense, coherent regions of behavior, and the height of a cell above the floor of its basin (its depth) measures how far it sits on a ridge between them. Rows on high ridges are never removed. This veto acts as a brake on how much traffic is cleared, and removing it degrades the model markedly.\\

\noindent\includegraphics[width=\linewidth, height=10cm]{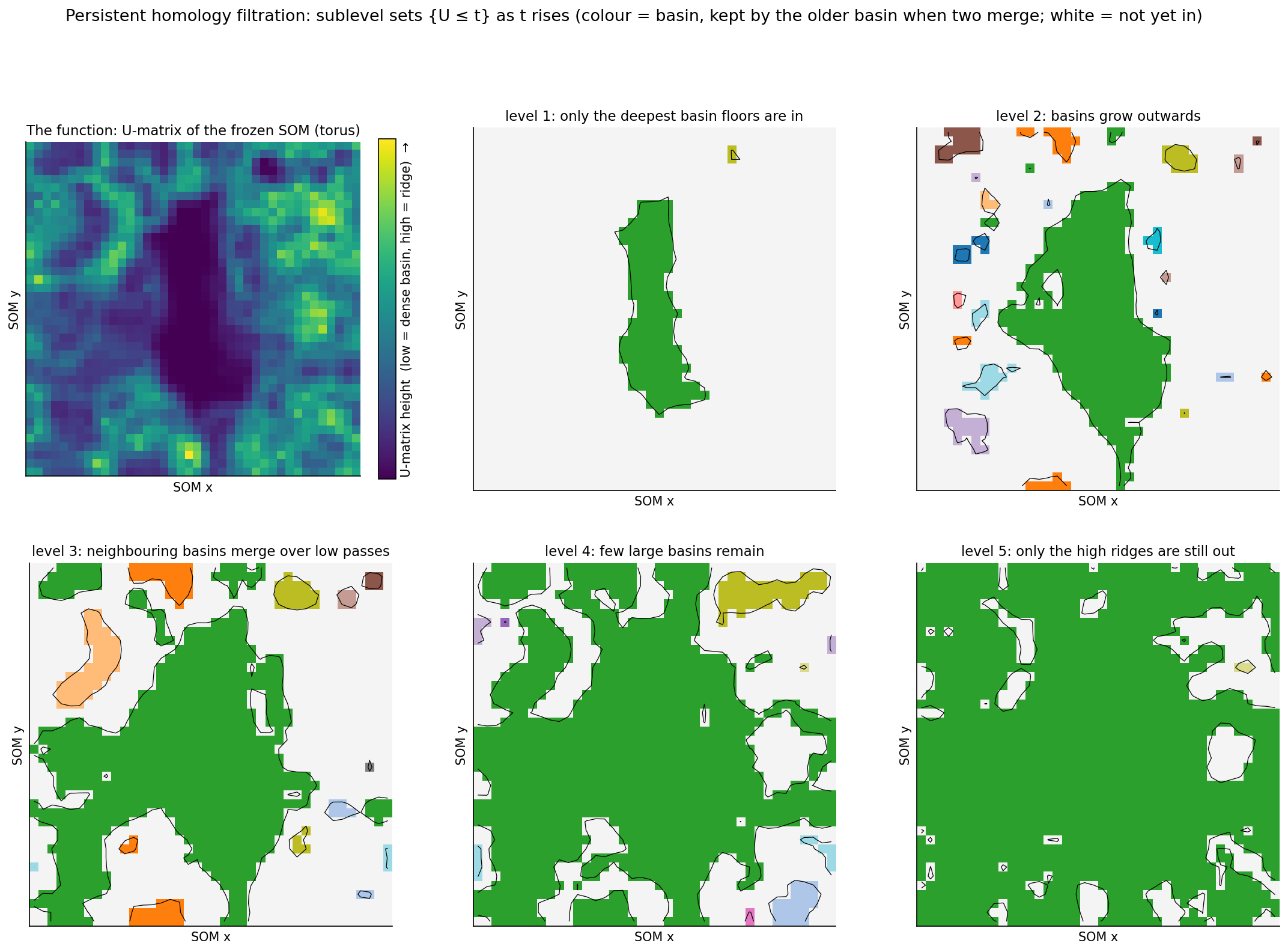}

\noindent\includegraphics[width=\linewidth, height=10cm]{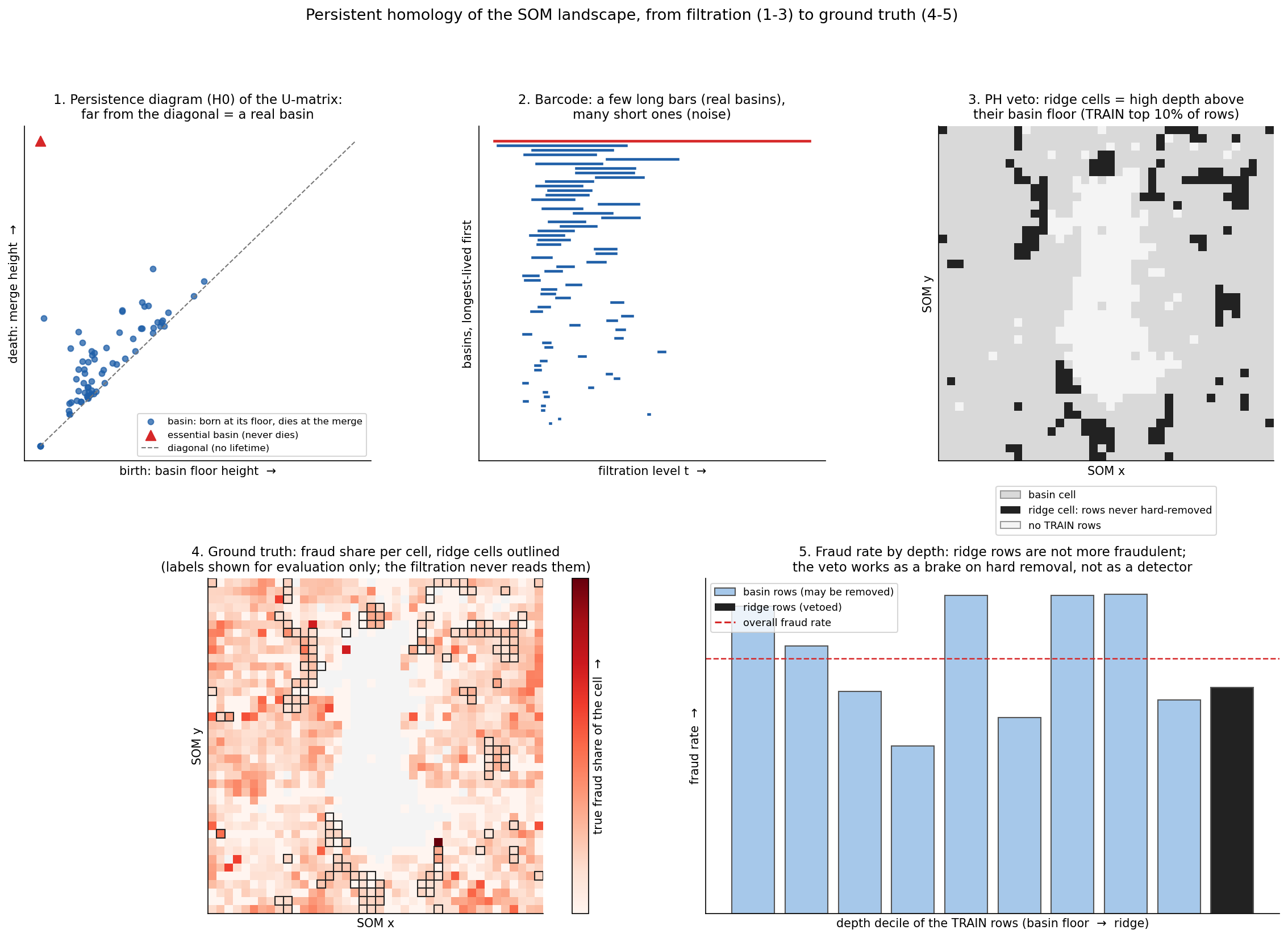}

\noindent Rows that are not removed but that the round's veto models flag as hot are \emph{set aside}: they skip the remaining iterations and go straight to the final classifier, carrying the index of the round that set them aside. All other rows proceed to the next iteration, where fresh SOMs are trained on what remains. The iterations use no supervised round classifier; labels enter only through the final classifier.

\subsection{The Final Classifier}

\noindent Every row that was not hard-removed, that is, those that survived all iterations in tandem with the set-aside hot rows, is scored by a single LightGBM gradient-boosting classifier \cite{LightGBM}. Chosen for fast inference and strong precision under our latency budget, its leaf-wise tree growth captures complex interactions with few nodes. Its inputs combine the topological history of the transaction with privacy-preserving descriptors of the transaction itself: the latent coordinates and the AE distance; the round history (how often the row was flagged hot, its last quantization error, how many rounds it survived, and the round that set it aside); the persistent-homology lookups of its SOM cell, from the frozen U-matrix and from vineyards across rounds \cite{Vineyards}. Raw columns from which quasi-identifiers are removed; and simple per-row features describing time, cents patterns, count ratios and frequency tables frozen on the training data. No input links transactions to a person.\\

\noindent\includegraphics[width=\linewidth, height=6cm]{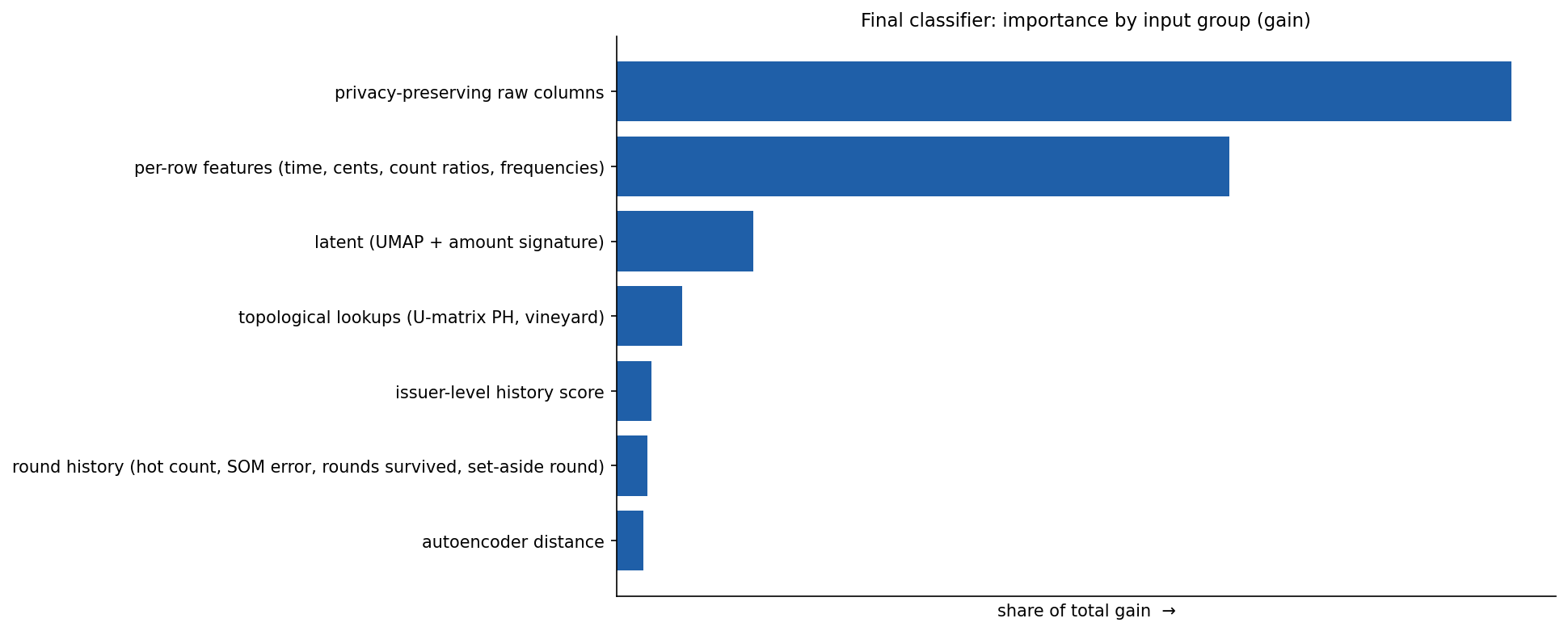}

\noindent The decision threshold is set by a false-positive budget rather than by a recall target. In deployment it is re-estimated as a quantile of the recent score distribution, so that the false-alarm rate stays on budget as traffic drifts. Model selection relies on forward-chained, out-of-time evaluation over successive blocks of the training period together with the validation period, repeated over several random seeds; the test period is reserved for the final report.\\

\subsection{Real-Time Detection of Fraud Rings}

\noindent Coordinated fraud rings reuse the same device across many cards in short bursts, and an individual transaction from a ring can look entirely ordinary to the main model. We therefore annex a lightweight submodel, \emph{ring mode}, to the pipeline; it never alters the main model and acts only on its final score. Each transaction's device fingerprint (device model, operating system, browser, screen resolution and two further anonymized identity fields, used only when complete) is reduced to a keyed cryptographic hash, and a short-lived store records, for every hash, how many of its transactions were later confirmed fraudulent or legitimate. Only labels that are already known when a transaction arrives are used, so the store respects the delay with which fraud is confirmed in practice. Because the store holds only hashes and counts, it can be shared between institutions without exchanging customer data, so that a ring confirmed at one bank is flagged at another.\\ 

\noindent We simulated this on top of the existing model by splitting the data into several independent institutions running the same model. Sharing helps, and increasingly so as the number of institutions grows, since the first confirmed hit of a ring then usually occurs elsewhere; the combined benefit equals that of a single institution holding everyone's history. The main limiting factor is speed: most ring activity occurs before any institution has confirmed the ring, so faster confirmation and near-real-time sharing of alerts matter more than any choice of identifier.

\section{Results and Conclusions}

\noindent The decision threshold is set by a false-positive budget rather than by a recall target. In deployment it is re-estimated as a quantile of the recent score distribution, so that the false-alarm rate stays on budget as traffic drifts. Model selection relies on forward-chained, out-of-time evaluation over successive blocks of the training period together with the validation period, repeated over several random seeds; the test period is reserved for the final report. The held-out test period is the final $20\%$ of the data in time: $118{,}108$ transactions, of which $4{,}064$ are fraudulent. Operating at a $0.15\%$ false-positive budget with the rolling threshold, the final model yielded $TP=1141$, $FN=2923$, $FP=190$, $TN=113854$, that is, $85.73\%$ precision and $28.08\%$ recall at a false-positive rate of $0.167\%$, for an F1 score of $0.4230$. These figures are end to end: fraud removed during the iterations counts as missed. With the annexed ring mode described below, recall rises to $28.57\%$ at a false-positive rate of $0.170\%$. For this architecture, a compiled single-transaction path on the author's consumer-grade AMD Ryzen 7 3700U machine takes $1.47$ ms at the median and $4.54$ ms at the 99th percentile from the latent representation, and about $2.5$ ms in a typical end-to-end evaluation including the UMAP serving transform.\\

\noindent Beyond traditional classification metrics, we evaluate the model using a weighted cost function: $C_{13} = 13\cdot \text{FP} + \text{FN}$, aligned with contemporary industry benchmarks established by the Merchant Risk Council \cite{MRC2024}. The $13$ weight is applied to account for the substantial economic attrition caused by false positive declines, where the immediate loss of a single fraudulent transaction is significantly outweighed by the long-term impact on Customer Lifetime Value following an erroneous rejection. Under this weighting a flag lowers $C_{13}$ only if its precision exceeds $13/14 \approx 92.9\%$. At the $0.15\%$ budget, where precision is $85.7\%$, $C_{13} = 5{,}393$, above the $4{,}064$ incurred by flagging nothing. Minimizing $C_{13}$ therefore calls for a far stricter operating point: a threshold chosen on the validation period to minimize $C_{13}$ flags $241$ test transactions at $95.4\%$ precision ($TP=230$, $FP=11$) and reaches $C_{13} = 3{,}977$, below the no-model baseline; the best value attainable on the test period in hindsight is $3{,}951$. The false-positive budget and the cost function thus describe two different operating regimes along the same ranking:

\noindent\includegraphics[width=\linewidth, height=6cm]{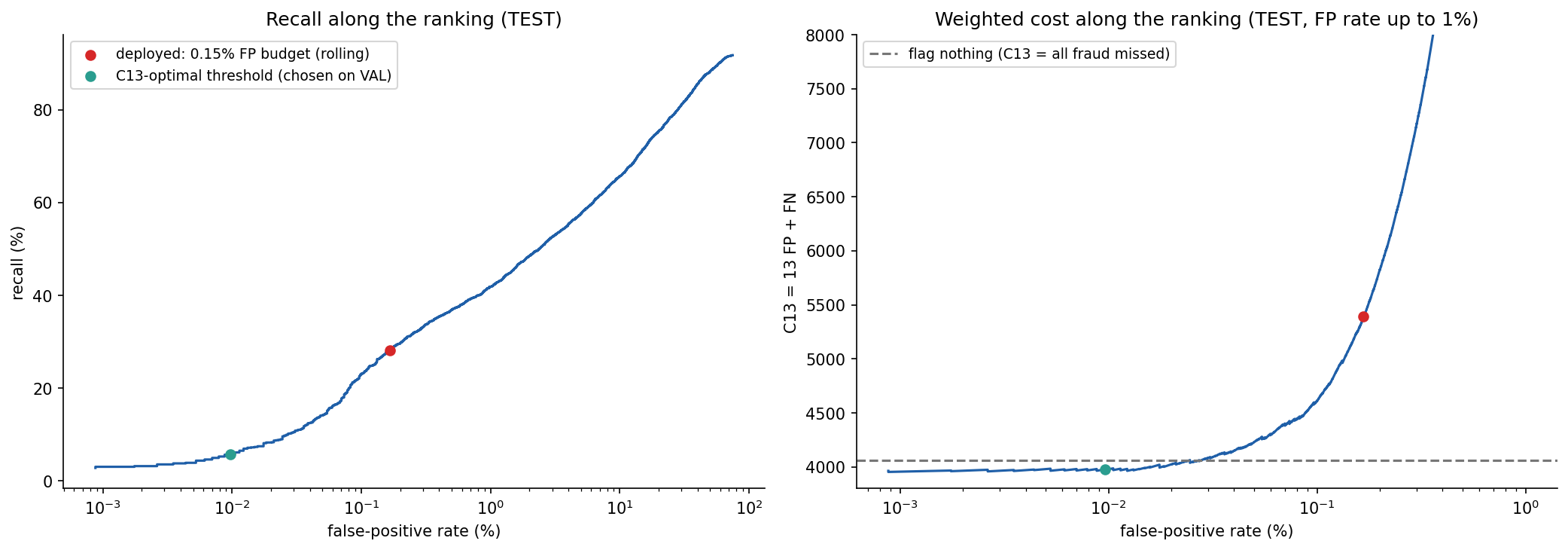}

\noindent For the subsequent fraud ring m,odel, every configuration is compared with the main model acting alone at exactly the same number of false alarms, and confidence intervals are obtained by bootstrapping over rings. The held-out test period contains 300 fraudulent transactions that belong to rings. Within a single institution, ring mode raises the share of ring fraud caught from $52.3\%$ to $56.7\%$ (from $157$ to $170$ transactions) and overall recall from $28.1\%$ to $28.6\%$, while the false-positive rate moves only from $0.167\%$ to $0.170\%$. Across institutions, with fraud confirmed one day after the transaction, sharing exact fingerprints raises ring recall by $6.7$ points ($95\%$ CI $2.2$--$11.8$) at a cost of about $1.5\%$ of all catches, whatever the number of institutions. That number changes is the network's share of the gain: on top of a bank's own history, sharing adds $1.4$ points with two institutions and $2.7$ points with ten. With ten institutions, $39$ ring transactions occur after another bank has already confirmed their ring; a bank acting alone catches $19$ of them, and $28$ with sharing. By contrast, $78\%$ of ring transactions occur before any institution has confirmed their ring ($89\%$ with a seven-day confirmation delay), which bounds what any sharing of confirmed fraud can achieve. When rings rotate their devices the network gain disappears, and neither coarser identifier keys nor proximity in the latent space recovers it.\\

\noindent We have demonstrated that an iterative topological sniping and filtering pipeline can effectively identify fraud within anonymized latent spaces at a latency of a few milliseconds per transaction. The annexed fraud-ring submodel confirms that the pipeline can host a privacy-preserving network layer: measured against each institution alone at an identical false-alarm count, sharing hashed device fingerprints catches significantly more ring fraud, and the network's share of that gain grows as more institutions participate. Its reach is bounded by how quickly fraud is confirmed rather than by the model, which makes near-real-time sharing of alerts the natural next step. Future research must evaluate whether this classifier maintains its efficacy across disparate vendor datasets and shifting temporal scales.\\

\noindent\textbf{Acknowledgments:}
\noindent The author would like to thank Amir Rozenfeld for his suggestions, encouragement, and good company.

\end{document}